\pdfoutput=1

\documentclass[11pt]{article}

\usepackage{acl}

\usepackage{times}
\usepackage{latexsym}
\usepackage{amssymb}
\usepackage{multirow}
\usepackage{booktabs}
\usepackage{hyperref}

\usepackage[T1]{fontenc}

\usepackage[utf8]{inputenc}
\usepackage{CJKutf8}

\usepackage{graphicx}
\graphicspath{ {./figures/} }

\usepackage{microtype}

\usepackage{xcolor}

\usepackage{bm}

\def\eqref#1{equation~\ref{#1}}

\def\1{\bm{1}}

\def\ve{{\bm{e}}}

\def\vu{{\bm{u}}}

\def\vw{{\bm{w}}}

\def\mE{{\bm{E}}}

\def\mU{{\bm{U}}}

\def\mW{{\bm{W}}}

\DeclareMathAlphabet{\mathsfit}{\encodingdefault}{\sfdefault}{m}{sl}
\SetMathAlphabet{\mathsfit}{bold}{\encodingdefault}{\sfdefault}{bx}{n}

\def\sG{{\mathbb{G}}}

\def\sU{{\mathbb{U}}}
\def\sV{{\mathbb{V}}}

\newcommand{\E}{\mathbb{E}}

\newcommand{\Var}{\mathrm{Var}}

\title{WECHSEL: Effective initialization of subword embeddings for cross-lingual transfer of monolingual language models}

\usepackage{array}

\newcommand{\eg}{e.\,g. }
\newcommand{\ie}{i.\,e. }
\newcommand{\wrt}{w.\,r.\,t. }
\newcommand{\cf}{c.\,f. }

\newcommand{\ourroberta}{WECHSEL-RoBERTa}
\newcommand{\ouraltroberta}{WECHSEL\textsubscript{TFR}-RoBERTa}
\newcommand{\randomembroberta}{TransInner-RoBERTa}
\newcommand{\fullrandomroberta}{FullRand-RoBERTa}
\newcommand{\ourgpt}{WECHSEL-GPT2}
\newcommand{\ouraltgpt}{WECHSEL\textsubscript{TFR}-GPT2}
\newcommand{\randomembgpt}{TransInner-GPT2}
\newcommand{\fullrandomgpt}{FullRand-GPT2}
\newcommand{\ourmethod}{WECHSEL}
\newcommand{\ouraltmethod}{WECHSEL\textsubscript{TFR}}
\newcommand{\randomembmethod}{TransInner}
\newcommand{\fullrandommethod}{FullRand}
\newcommand{\vocabulary}{$\sV$}
\newcommand{\wordembeddings}[1]{$\mW^{#1}$}
\newcommand{\subwordembeddings}[1]{$\mU^{#1}$}

\author{Benjamin Minixhofer{\normalfont \textsuperscript{1}} \and Fabian Paischer{\normalfont \textsuperscript{2,3}} \and Navid Rekabsaz{\normalfont \textsuperscript{1,3}}\\
\textsuperscript{1}Institute of Computational Perception, Johannes Kepler University Linz\\
\textsuperscript{2}Institute for Machine Learning, Johannes Kepler University Linz\\
\textsuperscript{3}ELLIS Unit Linz and LIT AI Lab\\
\texttt{\{benjamin.minixhofer, navid.rekabsaz\}@jku.at}\\
\texttt{paischer@ml.jku.at}\\
}

\begin{document}
\maketitle
\begin{abstract}

Large pretrained language models (LMs) have become the central building block of many NLP applications. Training these models requires ever more computational resources and most of the existing models are trained on English text only. It is exceedingly expensive to train these models in other languages. To alleviate this problem, we introduce a novel method -- called \ourmethod{} -- to efficiently and effectively transfer pretrained LMs to new languages. \ourmethod{} can be applied to any model which uses subword-based tokenization and learns an embedding for each subword. The tokenizer of the source model (in English) is replaced with a tokenizer in the target language and token embeddings are initialized such that they are semantically similar to the English tokens by utilizing multilingual static word embeddings covering English and the target language. We use \ourmethod{} to transfer the English RoBERTa and GPT-2 models to four languages (French, German, Chinese and Swahili). We also study the benefits of our method on very low-resource languages. \ourmethod{} improves over proposed methods for cross-lingual parameter transfer and outperforms models of comparable size trained from scratch with up to 64x less training effort. Our method makes training large language models for new languages more accessible and less damaging to the environment. We make our code and models publicly available.

\end{abstract}

\section{Introduction}

Large LMs based on the Transformer architecture~\cite{NIPS2017_3f5ee243} have become increasingly popular since GPT~\cite{radford2018improving} and BERT~\cite{devlin-etal-2019-bert} were introduced, prompting the creation of many large LMs pretrained on English text \cite{NEURIPS2019_dc6a7e65,clark2020electra,NEURIPS2020_d6f1dd03,ram-etal-2021-shot}. There is a tendency towards training larger and larger models~\cite{brown2020language,fedus2021switch} while the main focus is on the English language. Recent work has called attention to the costs associated with training increasingly large LMs, including environmental and financial cost~\cite{strubell-etal-2019-energy, bender2021dangers}. If training large LMs for English is already costly, it is prohibitively expensive to train new, similarly powerful models to cover other languages.

One approach to address this issue is creating massively multilingual models
~\cite{devlin-etal-2019-bert,conneau-etal-2020-unsupervised,xue-etal-2021-mt5} trained on a concatenation of texts in many different languages. These models show strong natural language understanding capabilities in a wide variety of languages, but suffer from what \citet{conneau-etal-2020-unsupervised} call the \textit{curse of multilinguality}: beyond a certain number of languages, overall performance decreases on monolingual as well as cross-lingual tasks. Consistent with this finding, \citet{nozza2020mask} observe that monolingual LMs often outperform massively multilingual models. This might be attributed to superior quality of monolingual tokenizers over their multilingual counterparts~\cite{rust-etal-2021-good}. It is thus desirable to train monolingual models in more languages. Training monolingual models in non-English languages is commonly done by training a new model with randomly initialized parameters~\cite{antoun-etal-2020-arabert, louis2020belgpt2,martin-etal-2020-camembert,rekabsaz2019regularization}. However, to train a model with capabilities comparable to that of an English model in this way, presumably a similar amount of compute to what was used to train the English model would be required.

To address this issue, we introduce \ourmethod{},\footnote{Word Embeddings Can Help initialize Subword Embeddings in a new Language.} a novel method to transfer monolingual language models to a new language. \ourmethod{} uses multilingual static word embeddings between the source language and the target language to initialize model parameters. \ourmethod{} first copies all inner (non-embedding) parameters of the English model, and exchanges the tokenizer with a tokenizer for the target language. Next, in contrast to prior work doing random initialization~\cite{de-vries-nissim-2021-good}, the token embeddings in the target language are initialized such that they are close to semantically similar English tokens by mapping multilingual static word embeddings to subword embeddings. The latter step is particularly important considering that token embeddings take up roughly 31\% of the parameters of RoBERTa~\cite{liu2019roberta} and roughly 33\% of the parameters of GPT2~\cite{radford2019language}. Intuitively, semantically transferring embeddings instead of randomly initializing one third of the model should result in improved performance. Our parameter transfer provides an effective initialization in the target language, requiring significantly fewer training steps to reach high performance than training from scratch. As multilingual static word embeddings are available for many languages \cite{bojanowski2017enriching}, \ourmethod{} is widely applicable.

We conduct our experiments on RoBERTa and GPT-2 as representative models of encoder and decoder language models, respectively. We transfer the English RoBERTa model to four languages (French, German, Chinese and Swahili), and the English GPT-2 model to the same four plus another four very low-resource languages (Sundanese, Scottish Gaelic, Uyghur and Malagasy). We evaluate the transferred RoBERTa models on Named Entity Recognition (NER), and Natural Language Inference (NLI) tasks in the respective languages. The transferred GPT-2 models are evaluated in terms of Language Modelling Perplexity (PPL) on a held-out set. We compare \ourmethod{} with randomly initialized models (denoted as \fullrandommethod{}), as well as the recently proposed \randomembmethod{} method which only transfers the inner (non-embedding) parameters~\cite{de-vries-nissim-2021-good}. All mentioned models are trained under the same conditions (around 4 days on a TPUv3-8). We also compare our model with models of comparable size trained from scratch under significantly larger training regimes, in particular CamemBERT~\cite{martin-etal-2020-camembert} (French), GBERT$_\mathrm{Base}$~\cite{chan-etal-2020-germans} (German), and BERT$_\mathrm{Base}$-Chinese~\cite{devlin-etal-2019-bert}.

Results show that models initialized with \ourmethod{} outperform randomly initialized models and models initialized with \randomembmethod{} across all languages and all tasks, for both RoBERTa and GPT-2. In addition, strong performance is reached at a fraction of the training steps of other methods. Our contribution is summarized as follows.

\begin{itemize}
  \item We propose \ourmethod{}, a novel method for transferring monolingual language models to a new language by utilizing multilingual static word embeddings between the source and the target language.
  \item We show effective transfer of RoBERTa and GPT-2 using \ourmethod{} to four and eight languages, respectively, achieved after minimal training effort.
  \item We release more effective GPT-2 and RoBERTa models than previously published non-English models, achieved under our more efficient training setting. Our code and models are publicly available at \href{https://github.com/cpjku/wechsel}{\nolinkurl{github.com/cpjku/wechsel}}. 
\end{itemize}







In the following, we review related work in Section~\ref{sec:related_work}. We introduce the \ourmethod{} method in Section~\ref{sec:methodology}, followed by explaining the experiment setup in Section~\ref{sec:experimental_design}. We show and discuss results in Section~\ref{sec:results}.

\section{Related Work}
\label{sec:related_work}

\paragraph{Large Language Models.} Training Language Models is usually done in a self-supervised manner \ie deriving labels from the training text instead of needing explicit annotations. One optimization objective is Masked Language Modelling \cite[MLM]{devlin-etal-2019-bert}, where randomly selected tokens in the input are replaced by a special \verb|[MASK]| token, and the task is to predict the original tokens. Another common objective is Causal Language Modelling (CLM), where the task is to predict the next token. These two objectives highlight a fundamental distinction between language models: models can be trained as encoders (e.g. with MLM) or as decoders (e.g. with CLM).

Instead of words, the vocabulary of recently proposed language models commonly consists of subwords~\cite{clark2020electra, liu2019roberta, devlin-etal-2019-bert}.



\paragraph{Multilingual representations.} 

There has been a significant amount of work in creating multilingual static word embeddings. A common method is learning embeddings from scratch using data in multiple languages~\cite{luong-etal-2015-bilingual,duong-etal-2016-learning}.
Alternatively, multilinguality can be achieved by aligning existing monolingual word embeddings using a bilingual dictionary, so that the resulting embeddings share the same semantic space~\cite{xing-etal-2015-normalized,joulin-etal-2018-loss}. Recent studies improve on this by reducing (or completely removing) the need for bilingual data~\cite{artetxe-etal-2017-learning,artetxe-etal-2018-robust,lample2018word}.

Beside static word embeddings, multilinguality is also well studied in the area of contextualized representations. One approach to learn multilingual contextualized representations is through training a model on a concatenation of corpora in different languages. Some models created based on this approach are mBERT~\cite{devlin-etal-2019-bert}, XLM-R~\cite{conneau-etal-2020-unsupervised} and mT5~\cite{xue-etal-2021-mt5}, trained on text in 104, 100, and 101 languages, respectively. As shown by \citet{pires-etal-2019-multilingual}, a multilingual model such as mBERT can enable cross-lingual transfer by using task-specific annotations in one language to fine-tune the model for evaluation in another language. Despite the benefits, recent studies outline a number of limitations of massively multilingual LMs. \citet{wu-dredze-2020-languages} empirically show that in mBERT 
``the 30\% languages with least pretraining resources perform worse than using no pretrained language model at all''. \citet{conneau-etal-2020-unsupervised} report that beyond a certain number of languages in the training data, the overall performance decreases on monolingual as well as cross-lingual tasks. These studies motivate our work on introducing an efficient approach for creating effective monolingual LMs for more languages.

\paragraph{Cross-lingual transfer of monolingual LMs.} Studies in this area can be divided into two categories:

\begin{itemize}
    \item \textbf{Bilingualization of a monolingual LM} is concerned with extending a model to a new language while preserving its capabilities in the original language. \citet{artetxe-etal-2020-cross} approach this problem by replacing the tokenizer and relearning the subword embeddings, while freezing other (non-embedding) parameters. Such a model becomes bilingual, since the initial tokenizer and embeddings can be used for tasks in the source language, while the new tokenizer and embeddings can be used for tasks in the target language. Thus, a model can be finetuned on annotated task data in the source language, and then zero-shot transferred to the target language. \citet{tran2020english} follow a similar approach, while instead of randomly initializing embeddings, they utilize static word embeddings to initialize embeddings in the target language close to semantically similar English tokens. They then continue training the model on an English text corpus as well as on the target language in order to preserve model capabilities in English.
    
    \item \textbf{Creating a new monolingual LM in the target language} is, in contrast, concerned with transferring a model from a source to a target language without the necessity to preserve its capabilities in the source language. \citet{zoph-etal-2016-transfer} and \citet{nguyen-chiang-2017-transfer} show that cross-lingually transferring a machine translation model can improve performance, especially for low-resource languages. \citet{zoph-etal-2016-transfer} use embeddings of random tokens in the original vocabulary to initialize token embeddings in the new vocabulary, while \citet{nguyen-chiang-2017-transfer} utilize vocabulary overlap between the source and target language. More recently, \citet{de-vries-nissim-2021-good} follow a similar approach to the one of \citet{artetxe-etal-2020-cross} for transferring a GPT-2 model to a new language. \citet{de-vries-nissim-2021-good} add an additional step, where they train the entire model for some amount of steps to allow adapting to the target language beyond the lexical level. We refer to the method of \citet{de-vries-nissim-2021-good} as \randomembmethod{} and consider it as a baseline in our experiments.
\end{itemize}

Our \ourmethod{} method belongs to the second category. \ourmethod{} can be seen as an extension to the method proposed by \citet{tran2020english} with the goal of creating a new monolingual LM instead of bilingualizing the LM. This allows removing the constraints imposed by the need to preserve the model's capabilities in the source language. In addition, we generalize the semantic subword mapping done by \citet{tran2020english} to consider an arbitrary number of semantically similar subword with an arbitrary temperature. We are the first to show that a cross-lingually transferred model can outperform monolingual models which have been trained extensively from scratch in the target language, while requiring substantially less computational resources.

\section{Methodology}
\label{sec:methodology}

To initialize the model in the target language, we copy the inner (non-embedding) parameters from the source model. Our goal, then, is given the tokenizer $T^s$ in the source language with vocabulary $\sU^s$, the corresponding token embeddings $\mE^s$, and a tokenizer $T^t$ in the target language with vocabulary $\sU^t$, to find a good initialization of the embeddings $\mE^t$ by using $\mE^s$. To this end, we use existing bilingual word embeddings enriched with subword information, containing a set of words and subword n-grams in the source and target language and their aligned vectors. We denote the set of words and n-grams in the source and target language as \vocabulary$^{s}$ and \vocabulary$^{t}$ respectively, and the aligned static embeddings as \wordembeddings{s} and \wordembeddings{t}. In Appendix~\ref{appendix:no_subword_information} we consider an alternative method if no subword information is available in the bilingual word embeddings.

\begin{figure}[t]
\hspace*{-0.15cm}\includegraphics[width=7.8cm]{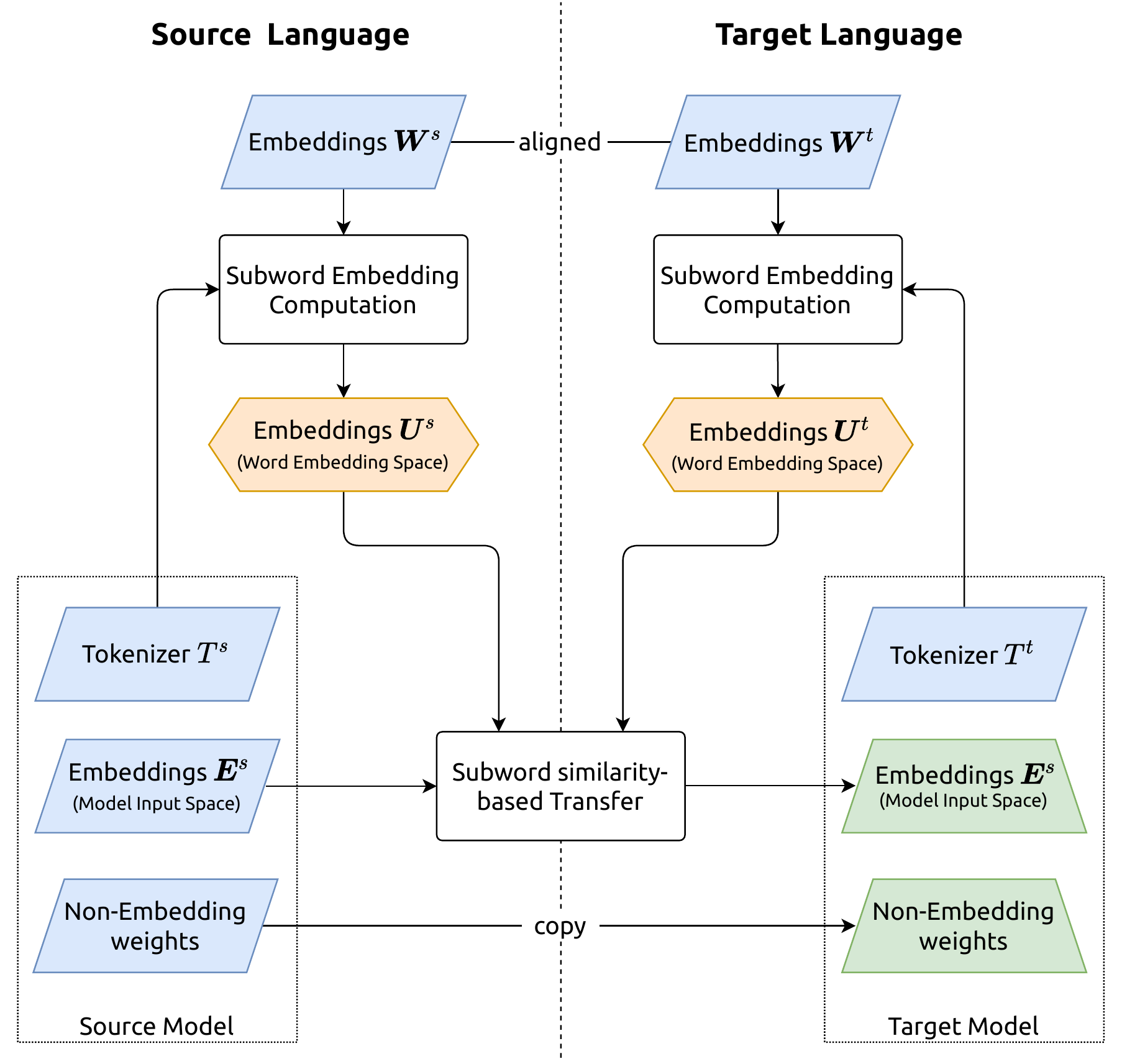}
\caption{Summary of our \ourmethod{} method. We show \textcolor{blue}{inputs}, \textcolor{orange}{intermediate results} and \textcolor{teal}{outputs}.}
\centering
\label{fig:flow}
\end{figure}

First, independently for both languages, we compute static subword embeddings for tokens in the tokenizer vocabulary in the same semantic space as the static word embeddings (Section \ref{section:subword-embedding-computation}). This results in subword embeddings \subwordembeddings{s} and \subwordembeddings{t} for the source and target language, respectively. Next, we use \subwordembeddings{s} and \subwordembeddings{t} to compute the semantic similarity of every subword in $\sU^s$ to every subword in $\sU^t$. Using these semantic similarities, we initialize the embeddings in $\mE^t$ through a convex combination of embeddings in $\mE^s$ (Section \ref{section:similarity-based}). By applying \ourmethod, the vectors of $\mE^t$ are in the same semantic space as $\mE^s$, where a subword in the target language is semantically similar to its counterpart(s) in the source language. These steps are summarized in Figure \ref{fig:flow} and explained in more detail in the following. 

\subsection{Subword Embedding Computation}
\label{section:subword-embedding-computation}

The process of mapping word embeddings to subword embeddings is done individually for the source and the target language. Given a tokenizer $T$ with vocabulary $\sU$ and embeddings \wordembeddings{}, the goal is to find subword embeddings \subwordembeddings{} for subwords in $\sU$ in the same semantic space as \wordembeddings{}. To this end, we decompose subwords in $\sU$ into n-grams and compute the embedding by taking the sum of the embeddings of all occuring n-grams, equivalent to how embeddings for out-of-vocabulary words are computed in fastText~\cite{bojanowski2017enriching}.

\[
\vu_{x} = \sum_{g \in \sG^{(x)}} \vw_{g}
\]
where $\sG^{(x)}$ is the set of n-grams occuring in the subword $x$ and $\vw_{g}$ is the embedding of the n-gram~$g$. Subwords in which no known n-gram occurs are initialized to zero.

\subsection{Subword similarity-based Transfer}
\label{section:similarity-based}
Applying the previous step to both source and target language results in the subword embeddings \subwordembeddings{s} and \subwordembeddings{t} over the subword vocabularies $\sU^s$ and $\sU^t$, respectively. Our aim is to leverage these embeddings to find an effective transformation from $\mE^s$ to $\mE^t$. We first compute the cosine similarity of every subword $x \in \sU^t$ to every subword $y \in \sU^s$, denoted as $s_{x,y}$.
\[
s_{x,y} = \frac{\vu^{t}_{x} {\vu^{s}_{y}}^T}{\|\vu^{t}_{x}\|\|\vu^{s}_{y}\|}
\]

We now exploit these similarities to initialize embeddings in $\mE^t$ by a convex combination of embeddings in $\mE^s$. In particular, each subword embedding in $\mE^t$ is defined as the weighted mean of the $k$ nearest embeddings in $\mE^s$ according to the similarity values. The weighting is done by a softmax of the similarities with temperature $\tau$.
\[
\ve_{x}^{t} = \frac{\sum_{y \in \mathcal{J}_x} {\exp{(s_{x,y}/\tau)} \cdot \ve_{y}^{s}}}{\sum_{y' \in \mathcal{J}_x} \exp{(s_{x,y'}/\tau)}}
\]
where $\mathcal{J}_x$ is the set of $k$ neighbouring subwords in the source language. Subword embeddings for which \subwordembeddings{t} is zero are initialized from a random normal distribution $\mathcal{N}(\E[\mE^s], \Var[\mE^s])$.

\section{Experiment Design}
\label{sec:experimental_design}

We evaluate our method by transferring the English RoBERTa~\cite{liu2019roberta} and the English GPT-2 model~\cite{radford2019language} to French, German, Chinese and Swahili. We refer to these languages as \textit{medium-resource languages}. In addition, we study the benefits of our method on four \textit{low-resource languages}, namely Sundanese, Scottish Gaelic, Uyghur and Malagasy.

We evaluate \ourroberta{} by fine-tuning on XNLI \cite{conneau2018xnli}, and on the balanced train-dev-test split of WikiANN \cite{rahimi-etal-2019-massively, pan-etal-2017-cross} to evaluate NLI and NER performance, respectively. The hyperparameters used for fine-tuning are reported in Appendix~\ref{appendix:hyperparameters}. GPT-2 is evaluated by Perplexity (PPL) on a held-out set from the same corpus on which the model was trained on. Due to the difficulty of extrinsic evaluation on low-resource languages, we only train GPT-2 models in these languages, and evaluate their performance intrinsically via Language Modelling Perplexity on a held-out set.
We use the pretrained models $\textrm{RoBERTa}_{\mathrm{Base}}$ with 125M parameters, and the small GPT-2 variant with 117M parameters provided by HuggingFace's Transformers~\cite{wolf-etal-2020-transformers} in all experiments. 

Since under limited training regimes such as ours, using a smaller corpus does not in general degrade performance~\cite{martin-etal-2020-camembert}, we use a subset of 4GiB from the OSCAR corpus for German, French and Chinese.
For the other languages, we use data from the CC-100 corpus \cite{conneau-etal-2020-unsupervised} which contains 1.6GiB, 0.1GiB, 0.1GiB, 0.4GiB and 0.2GiB for Swahili, Sundanese, Scottish Gaelic, Uyghur and Malagasy, respectively. To obtain aligned word embeddings between the source and the target language we use monolingual fastText word embeddings\footnote{https://fasttext.cc}
~\cite{bojanowski2017enriching}. We align these embeddings using the Orthogonal Procrustes method~\cite{schonemann1966generalized,artetxe-etal-2016-learning} with bilingual dictionaries from MUSE\footnote{https://github.com/facebookresearch/MUSE}~\cite{conneau2017word} for French, German and Chinese and a bilingual dictionary from FreeDict\footnote{https://freedict.org}~\cite{banski2009freedict} for Swahili. For the low-resource languages, we use bilingual dictionaries scraped from Wiktionary.\footnote{available at \href{https://github.com/cpjku/wechsel}{\nolinkurl{github.com/cpjku/wechsel}}}

We choose temperature $\tau=0.1$ and neighbors $k=10$ for \ourmethod{} by conducting a parameter search over a grid with varying values for $k$ and $\tau$ using linear probes (Appendix~\ref{appendix:grid_search}). We train tokenizers in the target languages using a vocabulary size of 50k tokens and byte-level BPE~\cite{radford2019language}. After applying \ourmethod{}, we continue training RoBERTa on the MLM objective and GPT-2 on the CLM objective. We compare against two baseline methods.
 \begin{itemize}
     \item \textbf{\randomembmethod:} Randomly initializing $\mE^t$ while transferring all other parameters from the English model as in \citet{de-vries-nissim-2021-good}. After training only embeddings for a fixed amount of steps while freezing other parameters, the entire model is trained for the remaining steps. In preliminary experiments reported in Appendix~\ref{appendix:choosing_baseline}, we compare the method by \citet{zoph-etal-2016-transfer} with \randomembmethod{}, observing superior performance of \randomembmethod{}, so we choose \randomembmethod{} as the baseline for cross-lingual transfer in all our experiments.
     
     \item \textbf{\fullrandommethod:} Training from scratch in the target language, as is commonly done when training BERT-like or GPT-like models in a new language \cite{antoun-etal-2020-arabert, louis2020belgpt2, chan-etal-2020-germans, martin-etal-2020-camembert}.
 \end{itemize}

\begin{table}[t]
\small
\begin{tabular}{lrr}
\hline
\textbf{Model} & \textbf{Tokens trained on} & \textbf{Factor}\\
\hline
\ourroberta & 65.5B & 1.0x\\
\randomembroberta & 65.5B & 1.0x\\
\fullrandomroberta & 65.5B & 1.0x\\
\hline
CamemBERT & 419.4B & 6.4x \\
GBERT$_{\mathrm{Base}}$ & 255.6B & 3.9x\\
BERT$_{\mathrm{Base}}$-Chinese & 131.1B & 2.0x\\
\end{tabular}
\caption{Tokens trained on in the target language between our models and previous monolingual models.}
\label{table:training_time}
\end{table}

\begin{table*}[t]
\centering
\small
\setlength\tabcolsep{3pt}
\aboverulesep=0ex
\belowrulesep=0ex
\begin{tabular}{llrrr|rrr|rrrrrr}
\toprule
\multirow{2}{*}{\textbf{Lang}} & \multirow{2}{*}{\textbf{Model}} & \multicolumn{3}{c|}{\textbf{Score@0}} & \multicolumn{3}{c|}{\textbf{Score@25k}} & \multicolumn{3}{c}{\textbf{Score@250k}} & \multicolumn{3}{c}{\textbf{Score (more training)}}\\
& & \textbf{NLI} & \textbf{NER} & \textbf{Avg} & \textbf{NLI} & \textbf{NER} & \textbf{Avg} & \textbf{NLI} & \textbf{NER} & \textbf{Avg} & \textbf{NLI} & \textbf{NER} & \textbf{Avg}\\
\midrule
\multirow{5}{*}{French} & \ourroberta & \underline{78.25} & \underline{86.93} & \underline{82.59} & \underline{81.63} & \underline{90.26} & \underline{85.95} & \underline{\textbf{82.43}} & \underline{\textbf{90.88}} & \underline{\textbf{86.65}} &- & - & -\\
& \randomembroberta & 60.86 & 69.57 & 65.21 &65.49 & 83.82 & 74.66 &81.75 & 90.34 & 86.04 &- & - & -\\
& \fullrandomroberta & 55.71 & 70.79 & 63.25 &69.02 & 84.24 & 76.63 &75.28 & 89.30 & 82.29 &- & - & -\\
& CamemBERT & - & - & - &- & - & - &- & - & - &80.88 & 90.26 & 85.57\\
& XLM-R$_{\mathrm{Base}}$ & - & - & - &- & - & - &- & - & - &79.25 & 89.48 & 84.37\\
\midrule
\multirow{5}{*}{German} & \ourroberta & \underline{75.64} & \underline{84.53} & \underline{80.08} & \underline{81.11} & \underline{89.05} & \underline{85.08} & \underline{\textbf{81.79}} & \underline{\textbf{89.72}} & \underline{\textbf{85.76}} &- & - & -\\
& \randomembroberta & 58.51 & 65.23 & 61.87 &64.78 & 82.05 & 73.42 &80.75 & 89.30 & 85.02 & - & - & -\\
& \fullrandomroberta & 54.82 & 66.84 & 60.83 &68.02 & 81.53 & 74.77 &75.48 & 88.36 & 81.92 &- & - & -\\
& GBERT$_{\mathrm{Base}}$ & - & - & - &- & - & - &- & - & - &78.64 & 89.46 & 84.05\\
& XLM-R$_{\mathrm{Base}}$ & - & - & - &- & - & - &- & - & - &78.58 & 88.76 & 83.67\\
\midrule
\multirow{5}{*}{Chinese} & \ourroberta & \underline{63.23} & \underline{72.79} & \underline{68.01} & \underline{77.19} & \underline{79.07} & \underline{78.13} & \underline{\textbf{78.32}} & \underline{80.55} & \underline{\textbf{79.44}} &- & - & -\\
& \randomembroberta & 46.95 & 69.06 & 58.01 &52.96 & 73.35 & 63.16 &76.99 & 80.00 & 78.49 &- & - & -\\
& \fullrandomroberta & 44.24 & 57.95 & 51.09 &58.34 & 64.84 & 61.59 &71.38 & 78.35 & 74.86 &- & - & -\\
& BERT$_{\mathrm{Base}}$-Chinese & - & - & - &- & - & - &- & - & - &76.55 & \textbf{82.05} & 79.30\\
& XLM-R$_{\mathrm{Base}}$ & - & - & - &- & - & - &- & - & - &76.41 & 78.36 & 77.38\\
\midrule
\multirow{4}{*}{Swahili} & \ourroberta & \underline{60.28} & \underline{74.38} & \underline{67.33} & \underline{73.87} & \underline{87.63} & \underline{80.75} & \underline{\textbf{75.05}} & \underline{\textbf{87.39}} & \underline{\textbf{81.22}} &- & - & -\\
& \randomembroberta & 54.67 & 64.46 & 59.56 &58.85 & 80.27 & 69.56 &74.10 & 87.05 & 80.57 &- & - & -\\
& \fullrandomroberta & 50.59 & 62.35 & 56.47 &63.79 & 83.49 & 73.64 &70.34 & 87.34 & 78.84 &- & - & -\\
& XLM-R$_{\mathrm{Base}}$ & - & - & - &- & - & - &- & - & - &69.18 & 87.37 & 78.28\\
\bottomrule
\end{tabular}
\caption{Results from fine-tuning RoBERTa models. We report accuracy for NLI on XNLI and micro F1 score for NER on WikiANN. Results are averaged over 3 runs. We report scores before training (\textbf{Score@0}), after 10\% of steps (\textbf{Score@25k}) and after training (\textbf{Score@250k}). We also report results from fine-tuning prior monolingual models and $\mathrm{XLM\textup{–}R}$ (\textbf{Score (more training)}), all trained on more tokens than our models. For each language, the best results in every column are indicated with underlines. The overall best results including the comparison with existing monolingual/multilingual models of comparable size are shown in bold.}
\label{table:roberta_results}
\end{table*}

All models are trained for 250k steps with the same hyperparameters across all languages (reported in Appendix \ref{appendix:hyperparameters}). Training one model takes around 4 days on a TPUv3-8. For \ourmethod{} and \fullrandommethod{} we use a learning rate (LR) schedule with linear warmup from zero to the peak LR for the first 10\% of steps, followed by a linear decay to zero. For \randomembmethod{}, we perform two warmup phases from zero to peak LR, once for the first 10\% of steps for training embeddings only, then again for the remaining steps while training the entire model.

In addition to the mentioned baselines trained under this setting, we compare the results of RoBERTa models with existing comparable models trained from scratch with more training effort. We consider the total number of tokens the model has encountered in the target language, computed as the product of batch size $\times$ sequence length $\times$ train steps (shown in Table \ref{table:training_time}) as a proxy for training effort. We evaluate the performance of CamemBERT~\cite{martin-etal-2020-camembert} (French), GBERT$_{\mathrm{Base}}$~\cite{chan-etal-2020-germans} (German), and BERT$_\mathrm{Base}$-Chinese~\cite{devlin-etal-2019-bert} as existing monolingual LMs,\footnote{To the best of our knowledge there is no monolingual model available for Swahili.} as well as XLM-R$_{\mathrm{Base}}$~\cite{artetxe-etal-2020-cross} as a high-performing multilingual LM.

\section{Results}
\label{sec:results}

We present our results on transferring RoBERTa and GPT-2 from English to other languages, followed by analyzing training behavior. In Appendix~\ref{appendix:closest_tokens}, we provide a qualitative assessment of how well subword tokens are mapped between the source and the target languages.

\subsection{Transferring RoBERTa}

Table~\ref{table:roberta_results} reports the evaluation results of RoBERTa. As shown, models initialized with \ourmethod{} outperform models trained from scratch and models initialized with \randomembmethod{} across all languages. Surprisingly, close relatedness of the source and target language is not necessary to achieve effective transfer, as \eg on NLI \ourmethod{} improves absolute accuracy by $7.15\%$, $6.31\%$, $6.94\%$ and $4.71\%$ over models trained from scratch for French, German, Chinese and Swahili, respectively.

\begin{figure*}[t]
\hspace*{-0.5cm}\includegraphics[width=16.5cm]{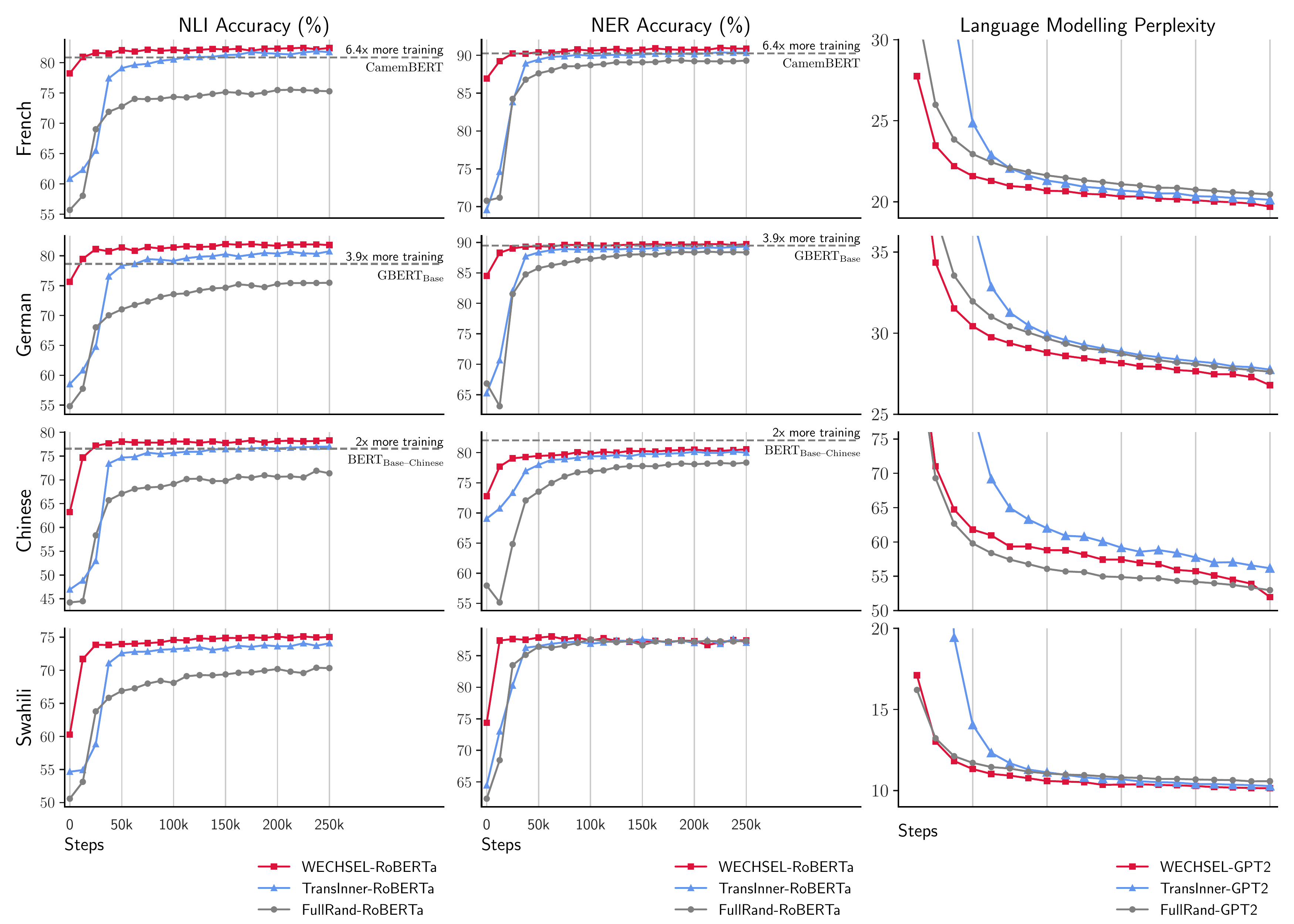}
\caption{Test scores over training steps from fine-tuning RoBERTa models on NLI (using XNLI) and NER (using WikiANN). Perplexity on the held-out set over training steps of GPT-2 models. We evaluate every 12.5k steps.}
\centering
\label{fig:scores}
\end{figure*}

We observe that our parameter transfer-based model consistently outperforms the previously released LMs on both monolingual and multilingual settings, while these models benefit from much larger training resources in terms of computation time and corpus size. In particular, the results show an improvement over XLM-R$_{\mathrm{Base}}$ by an average 3.54\% accuracy for NLI and 1.14\% micro F1 score for NER. For NLI, we improve over the prior monolingual models by $1.55\%$, $3.15\%$ and $1.77\%$ absolute accuracy for French, German and Chinese, respectively. For NER, we observe improvements over monolingual models with $0.62\%$ and $0.26\%$ absolute micro F1 score improvement for French and German, respectively. For Chinese, the monolingual model BERT$_\mathrm{Base}$-Chinese still outperforms our method by $1.5\%$ absolute micro F1 score. We suspect that the discrepancy between NLI and NER is due to the limited training corpus size (max. 4GiB), while a larger corpus can potentially improve NER as more named entities appear~\cite{martin-etal-2020-camembert}.

\begin{table}[t]
\centering
\small
\setlength\tabcolsep{1pt}
\begin{tabular}{llrrr}
\toprule
\textbf{Lang} & \textbf{Model} & \textbf{PPL@0} & \textbf{PPL@25k} & \textbf{PPL@250k}\\
\midrule
\multirow{3}{*}{French} & \ourgpt & \underline{1.7e+3} & \underline{23.47} & \underline{\textbf{19.71}}\\
& \randomembgpt & 1.4e+5 & 67.97 & 20.13\\
& \fullrandomgpt & 5.9e+4 & 25.99 & 20.47\\
\midrule
\multirow{3}{*}{German} & \ourgpt & \underline{3.7e+3} & \underline{34.35} & \underline{\textbf{26.80}}\\
& \randomembgpt & 1.5e+5 & 121.67 & 27.76\\
& \fullrandomgpt & 5.8e+4 & 37.29 & 27.63\\
\midrule
\multirow{3}{*}{Chinese} & \ourgpt & \underline{2.4e+4} & 71.02 & \underline{\textbf{51.97}}\\
& \randomembgpt & 1.5e+5 & 231.05 & 56.17\\
& \fullrandomgpt & 5.8e+4 & \underline{69.29} & 52.98\\
\midrule
\multirow{3}{*}{Swahili} & \ourgpt & 1.4e+5 & \underline{13.02} & \underline{\textbf{10.14}}\\
& \randomembgpt & 1.4e+5 & 42.95 & 10.28\\
& \fullrandomgpt & \underline{5.8e+4} & 13.22 & 10.58\\
\bottomrule
\end{tabular}
\caption{Results of training GPT2 models. We report Perplexity before training (\textbf{PPL@0}), after 10\% of steps (\textbf{PPL@25k}) and after training (\textbf{PPL@250k}).}
\label{table:gpt_results}
\end{table}

\begin{table}[t]
\centering
\small
\setlength\tabcolsep{1pt}
\begin{tabular}{llrrr}
\toprule
\textbf{Lang} & \textbf{Model} & \textbf{Best PPL}\\
\midrule
\multirow{3}{*}{Sundanese} & \ourgpt & \textbf{111.72}\\
& \randomembgpt & 151.86\\
& \fullrandomgpt & 149.46\\
\midrule
\multirow{3}{*}{Scottish Gaelic} & \ourgpt & \textbf{16.43}\\
& \randomembgpt & 18.62\\
& \fullrandomgpt & 19.53\\
\midrule
\multirow{3}{*}{Uyghur} & \ourgpt & \textbf{34.33}\\
& \randomembgpt & 39.06\\
& \fullrandomgpt & 42.82\\
\midrule
\multirow{3}{*}{Malagasy} & \ourgpt & \textbf{14.01}\\
& \randomembgpt & 14.85\\
& \fullrandomgpt & 15.93\\
\bottomrule
\end{tabular}
\caption{Results of training GPT2 models on low-resource languages. We report the best Perplexity on the held-out set, evaluated every 2.5k steps. See Figure~\ref{fig:low_resource_scores} for Perplexity throughout training.}
\label{table:low_resource_results}
\end{table}

The first two columns of Figure~\ref{fig:scores} show the performance of RoBERTa models on downstream tasks after each 12.5k training steps. Models initialized with \ourmethod{} reach high performance in significantly fewer steps than models initialized with \fullrandommethod{} or \randomembmethod{}. 

We expect \fullrandomroberta{} to approach performance of the respective prior monolingual models when trained on the same amount of tokens.\footnote{It would presumably be slightly worse because we restrict training corpus size to 4GiB.}
For French, \ourroberta{} outperforms CamemBERT after 10\% of training steps, reducing training effort by 64x. For German, \ourroberta{} outperforms GBERT$_\mathrm{Base}$ after 10\% of training steps, reducing training effort by 39x. For Chinese, \ourroberta{} outperforms BERT$_\mathrm{Base}$-Chinese on NLI, but does not 
outperform BERT$_\mathrm{Base}$-Chinese on NER.

\subsection{Transferring GPT-2}

\subsubsection{To Medium-Resource Languages}

Results on medium-resource languages are shown in Table~\ref{table:gpt_results}. Similar to the results for \ourroberta{}, the GPT-2 models trained with \ourmethod{} consistently outperform the models trained from scratch and with \randomembmethod{} across all languages. 

The rightmost column of Figure~\ref{fig:scores} depicts the performance of GPT-2 models after each 12.5k training steps. Comparing the results across all languages throughout training, we observe a stronger dependence on similarity of the source to the target language than for downstream tasks such as NLI or NER. In particular, for French and German, \ourmethod{} is consistently better than \randomembmethod{} and \fullrandommethod{} throughout the entire training, while for Chinese, a decrease in perplexity towards the end of training causes \ourmethod{} to surpass training from scratch.

\begin{figure}[h]
\includegraphics[width=7.9cm]{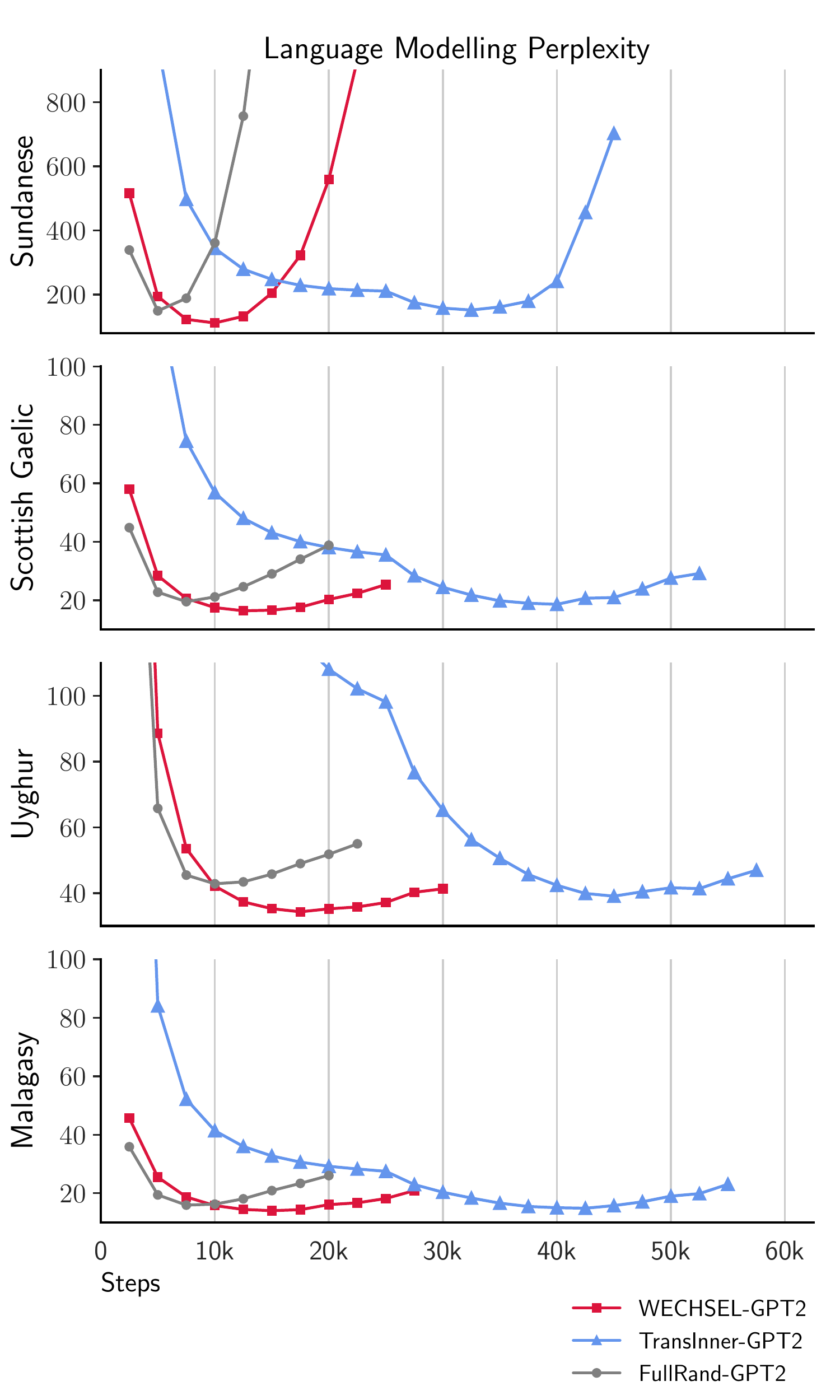}
\caption{Perplexity throughout training on low-resource languages. We evaluate every 2.5k steps and stop training if Perplexity on the held-out set does not improve for 10k steps.}
\centering
\label{fig:low_resource_scores}
\end{figure}

\subsubsection{To Low-Resource Languages}
Table~\ref{table:low_resource_results} reports the perplexity of Language Modelling on the low-resource languages. Again, we observe consistent improvements using \ourmethod{} on all languages. Furthermore, we find that the improvement from \ourmethod{} tends to increase as the amount of training data decreases by conducting a sensitivity analysis \wrt the amount of available training data (Appendix \ref{appendix:ablation}).

In Figure \ref{fig:low_resource_scores} we report the performance of the low-resource LMs on the held-out set throughout training. One difference of the low-resource models with the ones trained on medium-resource languages is that the low-resource LMs are prone to overfitting, and require appropriate model selection even in the early steps of training. Notably, \randomembgpt{} takes more steps to overfit since all non-embedding parameters are frozen for the first 25k steps (\cf Section \ref{sec:experimental_design}). 



\subsection{Is freezing necessary?}
\label{sec:freezing}

\begin{figure}[t]
\vspace*{-0.2cm}
\includegraphics[width=8cm]{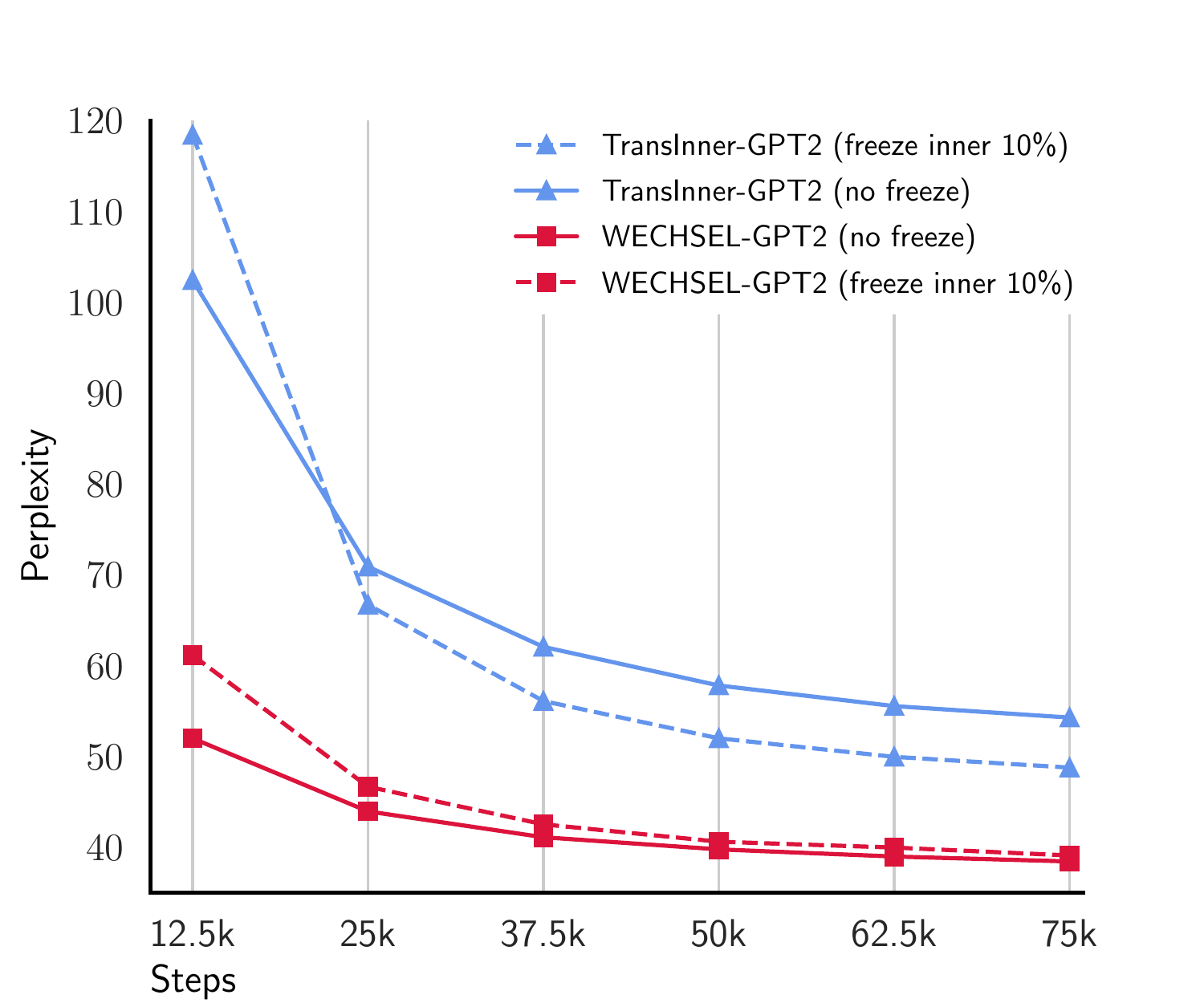}
\caption{Comparison of German GPT-2 models trained with \ourmethod{} and \randomembmethod{} between freezing non-embedding parameters at the start and not freezing any parameters.}
\centering
\label{fig:freezing}
\vspace*{-0.4cm}
\end{figure}

Previous work using the \randomembmethod{} method freezes non-embedding parameters for a fixed amount of steps before training the entire model~\cite{de-vries-nissim-2021-good}. This is done to prevent catastrophic forgetting at the beginning of training. To evaluate if freezing non-embedding parameters is still necessary with our method, we conduct an additional experiment. We train a German GPT-2 model with \ourmethod{} and a model with \randomembmethod{} without freezing any parameters, and the same models with freezing of non-embedding parameters for the first 10\% of steps. We match hyperparameters of the main experiments except training for 75k steps only. Based on the results shown in Figure \ref{fig:freezing}, we conclude that freezing is necessary when using \randomembmethod{}, but there is no need for freezing when using \ourmethod{}.

\section{Limitations and Potential Risks}

\vspace{-0.1cm}
\subsection{Limitations}

We conduct our experiments on up to eight languages, showing the benefits of our parameter transfer method to both medium- and low-resource languages. However, there are many more languages with diverse linguistic characteristics on which our \ourmethod{} method is not tested. This is a limitation forced by computational constraints, as we can not ascertain whether transfer to all other languages would result in similar improvements. In addition, our extrinsic evaluation is limited to two tasks (NLI and NER). While this choice is due to the limitations on the available collections in various languages, this evaluation does not necessarily provide a comprehensive view of language understanding tasks.

\vspace{-0.1cm}
\subsection{Risks}

It is well-known that existing LMs trained on English text encode societal biases~\cite{bolukbasi2016man,caliskan2017semantics,Rekabsaz0HH21}
 and stereotypes and using them in downstream tasks might lead to unfair treatment of various social groups~\cite{zerveas2022mitigating,krieg2022do,ganhoer2022mitigating,rekabsaz2021societal,melchiorre2021investigating,rekabsaz2020neural,elazar2018adversarial}. Since we propose a method to transfer the English LMs to new languages, it is highly probable that the existing biases are also transferred to the target LMs. We therefore advocate a conscious and responsible use of the transferred LMs in practice.

\section{Conclusion}

We introduce \ourmethod{}, an effective method to transfer monolingual language models to new languages. \ourmethod{} exploits multilingual static word embeddings to compute an effective initialization of subword embeddings in the target language. We conduct experiments by transferring RoBERTa and GPT-2 models from English to French, German, Chinese and Swahili, as well as English GPT-2 to four low-resource languages. The evaluation results show that the transferred RoBERTa and GPT-2 models are more efficient and effective than strong baselines, and consistently outperform prior monolingual models that have been trained for a significantly longer time. \ourmethod{} facilitates the creation of effective monolingual LMs for new languages with medium to low resources, particularly in computationally-limited settings. In addition, our work provides strong evidence towards the hypothesis by \citet{artetxe-etal-2020-cross} that deep monolingual language models learn abstractions that generalize across languages.

\section{Acknowledgments}

Research supported with Cloud TPUs from Google's TPU Research Cloud (TRC). We thank Andy Koh and Artus Krohn-Grimberghe for providing additional computational resources. The ELLIS Unit Linz, the LIT AI Lab, the Institute for Machine Learning, are supported by the Federal State Upper Austria. We thank the project INCONTROL-RL (FFG-881064). Research also supported in part by the NSF (IIS-1956221), the State of Upper Austria and the Austria's Federal Ministry of Education, Science, and Research through the project FAIRFLOW (LIT-2021-YOU-215).

\bibliography{anthology,custom}
\bibliographystyle{acl_natbib}

\appendix

\section{Grid search over $k$ and $\tau$}
\label{appendix:grid_search}

To choose number of neighbors $k$ and temperature $\tau$ for \ourmethod{} we conduct a grid search over linear probes of models with different initialization shown in Table \ref{table:grid_search}. For RoBERTa, we compute scores on NLI (using XNLI) and POS tagging (using the French, German and Chinese GSD corpora in Universal Dependencies) using linear probes of the last hidden state. We probe on NLI by taking a concatenation of the mean of all token representations in the premise with the mean of all token representations in the hypothesis. We probe on POS tagging by taking the mean of all token representations belonging to each word. For GPT2, we compute Language Modelling Perplexity on the held-out set also used to evaluate performance of the trained models.

\section{Hyperparameters}
\label{appendix:hyperparameters}

Hyperparameters used to fine-tune RoBERTa on downstream tasks are shown in Table \ref{table:finetuning_hyperparameters}. Hyperparameters used to train models in our main experiments are shown in Table \ref{table:hyperparameters}.

\begin{table}[ht]
\centering
\setlength\tabcolsep{2.2pt}
\begin{tabular}{lrr}
\hline
\textbf{Parameter} & \textbf{NLI} & \textbf{NER} \\
\hline
peak learning rate & 2e-5 & 2e-5 \\
batch size & 128 & 32 \\
sequence length & 128 & 128\\
Adam $\epsilon$ & 1e-8 & 1e-8\\
Adam $\beta_1$ & 0.9 & 0.9\\
Adam $\beta_2$ & 0.999 & 0.999\\
train epochs & 2 & 10\\
warmup & 10\% of steps & 10\% of steps\\
warmup schedule & linear & linear\\
LR decay & linear to zero & linear to zero\\

\hline
\end{tabular}
\caption{Hyperparameters used to fine-tune RoBERTa models on NLI (XNLI) and NER (WikiANN).}
\label{table:finetuning_hyperparameters}
\end{table}

\begin{table}[h]
\centering
\setlength\tabcolsep{2.2pt}
\begin{tabular}{lrr}
\hline
\textbf{Parameter} & \textbf{RoBERTa} & \textbf{GPT2} \\
\hline
peak learning rate & 1e-4 & 5e-4 \\
batch size & 512 & 512 \\
sequence length & 512 & 512\\
weight decay & 0.01 & 0.01\\
Adam $\epsilon$ & 1e-6 & 1e-6\\
Adam $\beta_1$ & 0.9 & 0.9\\
Adam $\beta_2$ & 0.98 & 0.98\\
train steps & 250k & 250k\\
\hline
\end{tabular}
\caption{Hyperparameters of the models transferred from RoBERTa and GPT2.}
\label{table:hyperparameters}
\end{table}

\begin{table}[h]
\centering
\small
\setlength\tabcolsep{2.5pt}
\begin{tabular}{l|lrr|rrr}
\hline
\multirow{2}{*}{\textbf{Lang}} & \multirow{2}{*}{\textbf{Model}} & \multirow{2}{*}{\textbf{$k$}} & \multirow{2}{*}{\textbf{$\tau$}} & \multicolumn{3}{c}{\textbf{Scores}}\\
& & & & \textbf{NLI} & \textbf{POS} & \textbf{LM}\\
\hline
\multirow{7}{*}{French} & \multirow{5}{*}{\ourmethod@0} & 1 & 1 & 58.4 & 85.2 & 2.5e+5\\
& & 10 & 0.1 & 59.8 & 86.8 & 2.0e+5\\
& & 10 & 1   & 58.3 & 84.4 & 4.8e+5\\
& & 50 & 0.1 & 57.2 & 83.6 & 3.1e+6\\
& & 50 & 1   & 54.0 & 81.6 & 1.8e+7\\
& \fullrandommethod@0 & - & - & 46.3 & 60.6 & 5.7e+6\\
& CamemBERT & - & - & 63.5 & 93.6 & - \\
\hline
\multirow{7}{*}{German} & \multirow{5}{*}{\ourmethod@0} & 1 & 1 & 55.8 & 72.7 &  6e+5\\
& & 10 & 0.1 & 58.9 & 76.0 & 4.2e+5\\
& & 10 & 1   & 57.5 & 75.4 & 8.3e+6\\
& & 50 & 0.1 & 55.4 & 75.4 & 1.0e+7\\
& & 50 & 1   & 53.6 & 69.5 & 5.9e+7\\
& \fullrandommethod@0 & - & - & 44.5 & 49.1 & 6.2e+6\\
& GBERT$_{\mathrm{Base}}$ & - & - & 63.2 & 81.4 & -\\
\hline
\multirow{7}{*}{Chinese} & \multirow{5}{*}{\ourmethod@0} & 1 & 1 & 47.4 & 75.4 &  2.7e+6\\
& & 10 & 0.1 & 48.0 & 80.7 & 2.6e+6\\
& & 10 & 1   & 48.3 & 80.3 & 3.1e+6\\
& & 50 & 0.1 & 48.3 & 77.8 & 3.7e+7\\
& & 50 & 1   & 47.9 & 76.5 & 8.6e+7\\
& \fullrandommethod@0 & - & - & 37.5 & 53.7 & 5.8e+6\\
& BERT$_{\mathrm{Base}}$-Chinese & - & - & 61.9 & 91.9 & -\\
\end{tabular}
\caption{Grid search over the temperature $\tau$ and number of most similar tokens $k$ parameters of \ourmethod{}.}
\label{table:grid_search}
\end{table}

\begin{table}[h]
\small
\begin{tabular}{l|rr}
\hline
\textbf{Lang} & \textbf{Target Token} & \textbf{Closest English Token}\\
\hline
\multirow{10}{*}{French} & héritage & legacy\\
& tremp & soaked\\
& épiscop & bishop\\
& scandaleux & udicrous\\
& vertig & astonishing\\
& enregistrer & rec\\
& sucrés & sweets\\
& Emmanuel & Emmanuel\\
& entourage & confid\\
& secrétariat & ariat\\
\hline
\multirow{10}{*}{German} & machen & ize\\
& mit & with\\
& Sprichwort & proverb\\
& erischen & Austrian\\
& minuten & utes\\
& Haustechnik & umbing\\
& dringen & urgent\\
& verfeinern & refine\\
& umgebung & vironments\\
& ternehmen & irms\\
\hline
\multirow{10}{*}{Chinese} & \begin{CJK*}{UTF8}{gbsn}到处\end{CJK*} & everywhere\\
& \begin{CJK*}{UTF8}{gbsn}巧合\end{CJK*} & coinc\\
& \begin{CJK*}{UTF8}{gbsn}第三\end{CJK*} & third\\
& \begin{CJK*}{UTF8}{gbsn}杂交\end{CJK*} & recomb\\
& \begin{CJK*}{UTF8}{gbsn}利来\end{CJK*} & chnology\\
& \begin{CJK*}{UTF8}{gbsn}政务\end{CJK*} & Govern\\
& \begin{CJK*}{UTF8}{gbsn}石\end{CJK*} & stone\\
& \begin{CJK*}{UTF8}{gbsn}喊麦\end{CJK*} & sing\\
& \begin{CJK*}{UTF8}{gbsn}中海\end{CJK*} & iterranean\\
& \begin{CJK*}{UTF8}{gbsn}张某\end{CJK*} & defendant\\
\hline
\multirow{10}{*}{Swahili} & shirikishe & ive\\
& Harusi & Marriage\\
& pesile & ery\\
& tihani & graduate\\
& changi & ool\\
& kuugua & ingestion\\
& kuzidi & acclaim\\
& vipigo & Trouble\\
& dhamiri & conscience\\
& aliposimama & Slowly\\
\end{tabular}
\caption{Samples of tokens in each language and the corresponding closest tokens from the English vocabulary according to \ourmethod{}.}
\label{table:closest_tokens}
\end{table}

\section{Qualitative subword correspondence}
\label{appendix:closest_tokens}

We show a small random sample of tokens in the target language and their closest English token (according to \ourmethod{}) in Table \ref{table:closest_tokens}.

\section{Using Word Embeddings without subword information}
\label{appendix:no_subword_information}

\begin{figure*}[t]
\includegraphics[width=16cm]{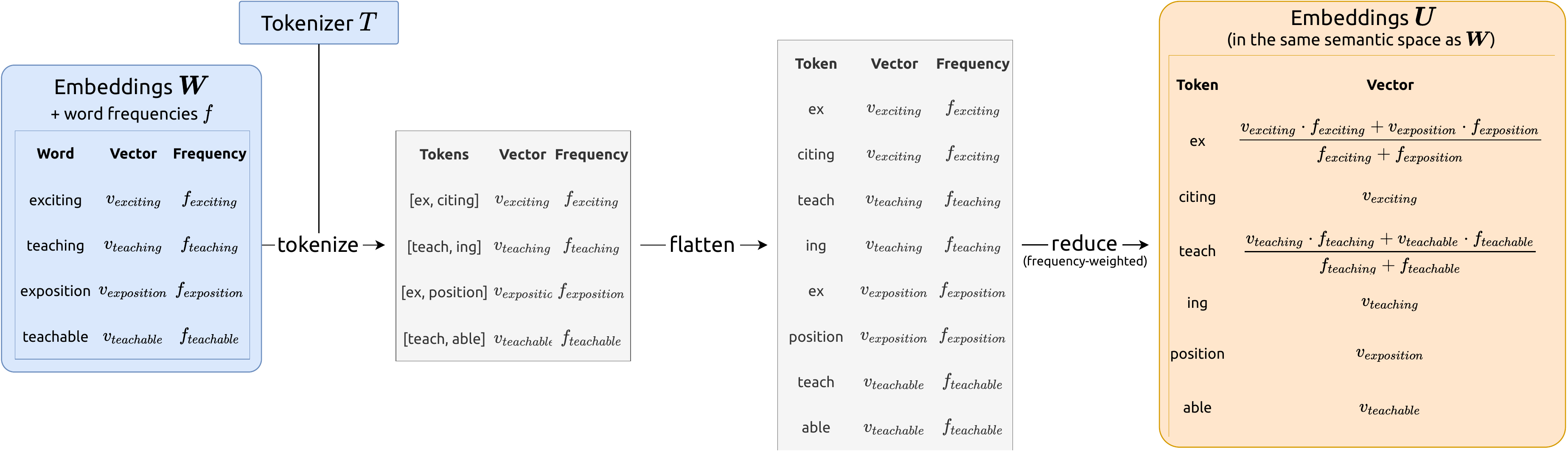}
\caption{\ouraltmethod{}, an alternative subword embedding computation method. First, \textbf{tokenize} all words in the word embeddings. Then \textbf{flatten} the result by assigning the embeddings of the words in which it occured and their word frequencies to each subword. Finally, \textbf{reduce} the embeddings assigned to each subword by taking their mean, weighted by word frequency.}
\centering
\label{fig:wordtosubword}
\end{figure*}

As an alternative to n-gram decomposition, we introduce a method for mapping word embeddings to subword embeddings without using any subword information (shown in Figure \ref{fig:wordtosubword}). For this method, we require word frequency information in addition to the word embeddings. We apply the tokenizer $T$ to every word $v$ in \vocabulary{} resulting in a set of subwords for each word. We define \vocabulary$^{(x)}$ as the set of words containing the subword $x$ when tokenized. The embedding $\vu_{x}$ of the subword $x$ is then defined as the average of the embeddings of words in \vocabulary$^{(x)}$, weighted by the word frequencies.

\[
\vu_{x} = \frac{\sum_{v \in \sV^{(x)}}{\vw_{v} \cdot f_{v}}}{\sum_{v \in \sV^{(x)}} f_{v}}
\]
where $\vw_{v}$ is the embedding and $f_v$ is the frequency of word $v$.\\
We call this variant of our method \ouraltmethod{}. We evaluate \ouraltmethod{} by training the same models as for \ourmethod{}. Results are shown in Table \ref{table:gpt2_results_tfr} for GPT2 and in Table \ref{table:roberta_results_tfr} for RoBERTa. We find that, on average, performance is on par with \ourmethod{}.

\begin{table*}[t]
\centering
\small
\setlength\tabcolsep{1pt}
\scalebox{0.95}{
\begin{tabular}{llrrr}
\toprule
\textbf{Lang} & \textbf{Model} & \textbf{PPL@0} & \textbf{PPL@25k} & \textbf{PPL@250k}\\
\midrule
\multirow{2}{*}{French} & \ourgpt & \underline{1.7e+3} & 23.47 & 19.71\\
& \ouraltgpt & 2.3e+3 & \underline{23.45} & \underline{\textbf{19.70}}\\
\midrule
\multirow{2}{*}{German} & \ourgpt & \underline{3.7e+3} & \underline{34.35} & \underline{\textbf{26.80}}\\
& \ouraltgpt & 5.0e+3 & 34.46 & 26.82\\
\midrule
\multirow{2}{*}{Chinese} & \ourgpt & \underline{2.4e+4} & \underline{71.02} & \underline{\textbf{51.97}}\\
& \ouraltgpt & 2.5e+4 & 72.11 & 52.07\\
\midrule
\multirow{2}{*}{Swahili} & \ourgpt & \underline{1.4e+5} & \underline{13.02} & 10.14\\
& \ouraltgpt & 1.5e+5 & 13.03 & \underline{\textbf{10.06}}\\
\bottomrule
\end{tabular}
}
\caption{Results of training \ouraltmethod{} GPT2 models. We report Perplexity before training (\textbf{PPL@0}), after 10\% of steps (\textbf{PPL@25k}) and after training (\textbf{PPL@250k}).}
\label{table:gpt2_results_tfr}
\end{table*}

\begin{table*}[t]
\centering
\setlength\tabcolsep{3pt}
\aboverulesep=0ex
\belowrulesep=0ex
\begin{tabular}{llrrr|rrr|rrrrrr}
\toprule
\multirow{2}{*}{\textbf{Lang}} & \multirow{2}{*}{\textbf{Model}} & \multicolumn{3}{c|}{\textbf{Score@0}} & \multicolumn{3}{c|}{\textbf{Score@25k}} & \multicolumn{3}{c}{\textbf{Score@250k}}\\
& & \textbf{NLI} & \textbf{NER} & \textbf{Avg} & \textbf{NLI} & \textbf{NER} & \textbf{Avg} & \textbf{NLI} & \textbf{NER} & \textbf{Avg}\\
\midrule
\multirow{2}{*}{French} & \ourroberta & \underline{78.25} & 86.93 & 82.59 & 81.63 & \underline{90.26} & 85.95 & 82.43 & \textbf{\underline{90.88}} & 86.65\\
 & \ouraltroberta & \underline{78.25} & \underline{87.43} & \underline{82.84} & \underline{81.86} & 90.07 & \underline{85.96} & \textbf{\underline{82.55}} & 90.80 & \textbf{\underline{86.68}}\\
\midrule
\multirow{2}{*}{German} & \ourroberta & 75.64 & 84.53 & 80.08 & \underline{81.11} & 89.05 & \underline{85.08} & 81.79 & \textbf{\underline{89.72}} & 85.76\\
& \ouraltroberta & \underline{77.00} & \underline{84.70} & \underline{80.85} & 80.71 & \underline{89.09} & 84.90 & \textbf{\underline{82.04}} & \textbf{\underline{89.72}} & \textbf{\underline{85.88}}\\
\midrule
\multirow{2}{*}{Chinese} & \ourroberta & \underline{63.23} & 72.79 & \underline{68.01} & \underline{77.19} & \underline{79.07} & \underline{78.13} & \textbf{\underline{78.32}} & 80.55 & \textbf{\underline{79.44}}\\
& \ouraltroberta & 62.75 & \underline{72.87} & 67.81 & 77.07 & 78.03 & 77.55 & 77.99 & \textbf{\underline{80.65}} & 79.32\\
\midrule
\multirow{2}{*}{Swahili} & \ourroberta & \underline{60.28} & 74.38 & 67.33 & 73.87 & 87.63 & 80.75 & \textbf{\underline{75.05}} & 87.39 & \textbf{\underline{81.22}}\\
& \ouraltroberta & 60.14 & \underline{75.42} & \underline{67.78} & \underline{74.04} & \underline{87.79} & \underline{80.92} & 74.58 & \textbf{\underline{87.66}} & 81.12\\
\bottomrule
\end{tabular}
\caption{Results from fine-tuning \ouraltroberta{} models. Results shown equivalently as in Table~\ref{table:roberta_results}.}
\label{table:roberta_results_tfr}
\end{table*}

\begin{figure}[t]
\includegraphics[width=8cm]{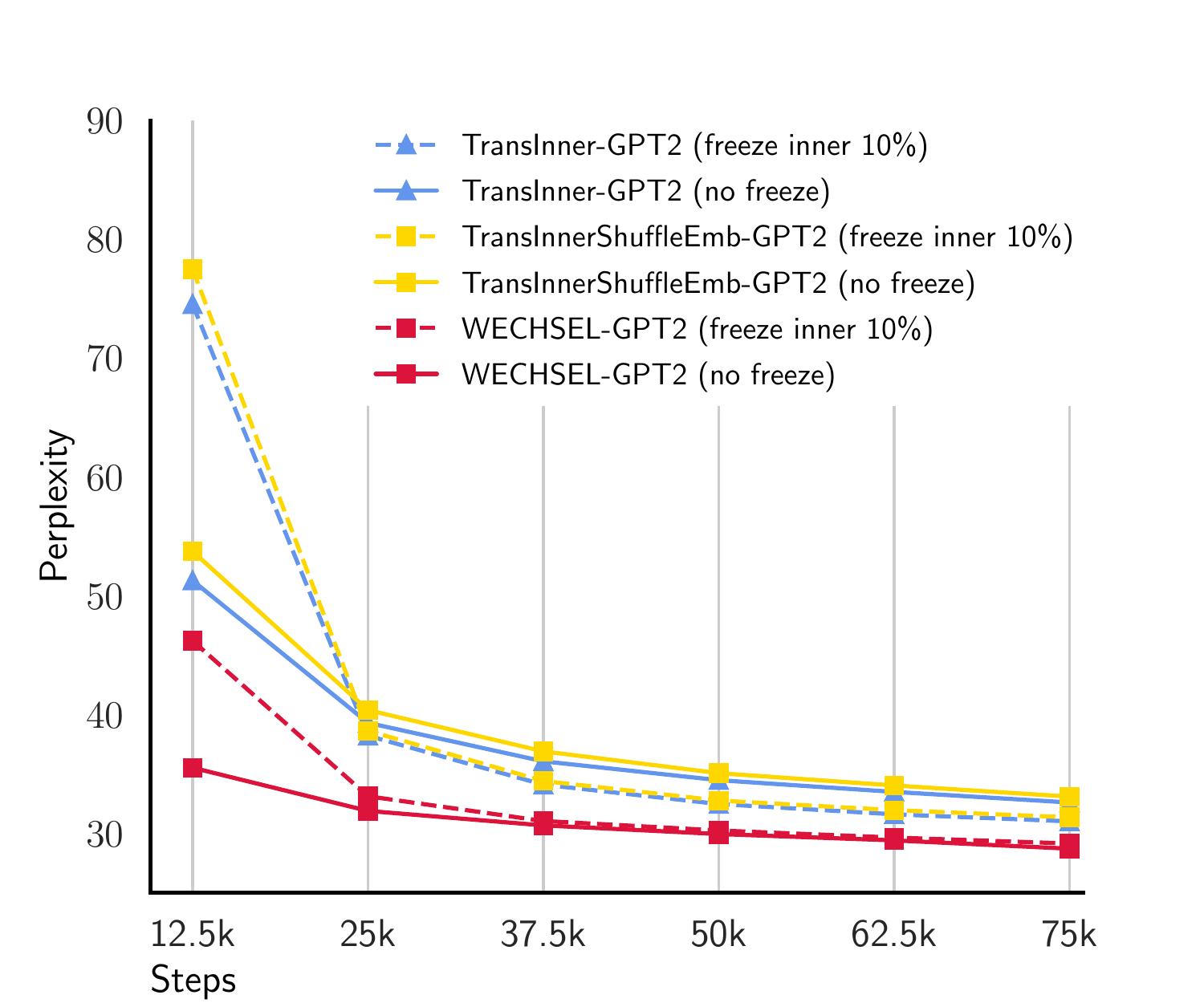}
\caption{Comparison of German GPT-2 models trained with \ourmethod{}, \randomembmethod{} and TransInnerShuffleEmb between freezing non-embedding parameters at the start and not freezing any parameters.}
\centering
\label{fig:choosing_baseline}
\end{figure}

\section{Choosing a transfer baseline}
\label{appendix:choosing_baseline}

We consider two baseline methods to transfer models to a new language without using any language-specific information. One method copies non-embedding parameters to the target language and initalizes embeddings from a random normal distribution as done by \citet{de-vries-nissim-2021-good}. We refer to this method as \randomembmethod{}. Another option copies non-embedding parameters and assigns the embedding of a random token in the source language to each embedding in the target language (effectively "shuffling" the embeddings) as done by \citet{zoph-etal-2016-transfer} and \citet{nguyen-chiang-2017-transfer}. We refer to this method as TransInnerShuffleEmb. We evaluate these two methods using a setup equivalent to the experiments in Section \ref{sec:freezing} and find that \randomembmethod{} performs slightly better than TransInnerShuffleEmb (Figure \ref{fig:choosing_baseline}), so we use \randomembmethod{} for subsequent experiments.

\vspace{1cm}

\section{Sensitivity Analysis \wrt training data size}
\label{appendix:ablation}

Evaluating on languages with different amounts of available data only indirectly measures the effect of training data size on \ourmethod{} since other factors (e.g. language similarity to English) are also involved. We conduct a sensitivity analysis to make the relation to the amount of training data explicit (Table \ref{table:ablation}). Due to computational constraints we only do this for French. We find that the improvement from \ourmethod{} increases as the amount of training data decreases. In addition, we find that using fastText embeddings trained on less data deteriorates performance, but still leaves a clear margin to \randomembmethod{} and \fullrandommethod{}.

\begin{table*}[t]
\centering
\setlength\tabcolsep{2pt}
\begin{tabular}{llrrr}
\toprule
& \multicolumn{4}{c}{\textbf{Best PPL}}\\
\textbf{Model \hspace*{\fill} Subsample Size} & 16MiB & 64MiB & 256MiB & 1024MiB\\
\midrule
\ourgpt{} (original fastText embeddings) & 78.33 & 44.75 & 31.63 & 24.66\\
\midrule
\ourgpt{} (fastText embeddings trained on subsample) & \underline{97.42} & \underline{49.50} & \underline{32.88} & \underline{24.75}\\
\fullrandomgpt{} & 281.46 & 83.43 & 43.08 & 27.09\\
\randomembgpt{} & 216.37 & 77.71 & 35.27 & 25.15\\
\bottomrule
\end{tabular}
\caption{Sensitivity Analysis \wrt the amount of training data on transfer to French. We train models on random subsamples of 16MiB, 64MiB, 256MiB and 1024MiB of the original training data, and evaluate on the same held-out set. For \ourgpt{}, we train two models. One using the original, publicly available fastText embeddings trained on Common Crawl data. The other using fastText embeddings trained only on the corresponding subsample of text.}
\label{table:ablation}
\end{table*}

\end{document}